# Development and Validation of a Low-Cost Imaging System for Seedling Germination Kinetics through Time-Cumulative Analysis

Torrente M.[1], Follador A.[1], Calcante A.[1], Casati P.[1], Oberti R.[1]

[1] DISAA - Department of Agricultural and Environmental Sciences, Università degli Studi di Milano - Via G. Celoria 2 - 20133 Milan, ITALY.

* roberto.oberti@unimi.it

## Abstract

This study investigates the effects of *Rhizoctonia solani* inoculation on the germination and early development of lettuce (*Lactuca sativa L.*) seeds using a low-cost, image-based monitoring system. Multiple affordable cameras were deployed to continuously capture images of the germination process in both infected and control groups. The primary objective was to assess the impact of the pathogen on seedling vigor by analyzing germination dynamics and growth patterns over time.

To achieve this, a novel image analysis pipeline was developed using MATLAB. The algorithm integrates both morphological and spatial features to accurately identify and quantify individual seedlings, even under complex conditions where traditional image segmentation fails. A key innovation of the method lies in its temporal integration: each analysis step considers not only the current morphological status of segmented clusters, but also their developmental history across prior time points. This cumulative approach enables robust discrimination of individual seedlings, especially during later growth stages when overlapping cotyledons and stems significantly hinder object separation.

The proposed method demonstrated high accuracy in seedling counting and vigor assessment, even in challenging scenarios characterized by dense and intertwined growth. Results confirm that *R. solani* infection significantly reduces germination rates and early seedling vigor. The study also validates the feasibility of combining low-cost imaging hardware with advanced computational tools to obtain phenotyping data in a non-destructive and scalable manner.

The imaging method based on temporal integration enabled accurate quantification of germinated seeds and precise determination of seedling emergence timing. This approach proved particularly effective in later stages of the experiment, where conventional segmentation techniques failed due to overlapping or intertwined seedlings, making accurate counting otherwise unfeasible. The method achieved a coefficient of determination ($R^2$) of 0.98 and a root mean square error (RMSE) of 1.12, demonstrating its robustness and reliability.

## 1 Introduction

The fresh-cut industry represents the most rapidly expanding sector within the fruit and vegetable market (Gullino *et al.,* 2019; Losio *et al.,* 2015). Lettuce (*Lactuca sativa L.*), a member of the Asteraceae family, is one of the most widely consumed salad crops globally. The demand for these products is notably high among consumers, driven by their convenience, quality, and the increasing awareness of the relationship between health and the consumption of fresh vegetables (Tomasi *et al.*, 2015). Its broad adaptability and widespread cultivation make it vulnerable to a variety of diseases caused by fungi, bacteria, viruses, and nematodes (Raid & Sandoya-Miranda, 2024). Among all these pathogens, *Rhizoctonia solani* is recognized as a highly destructive plant pathogen with a broad host range and is also a significant soil-borne pathogen (Verwaaijen *et al.*, 2017; Gonzalez Garcia *et al.*, 2006) in protected and open-fields lettuce crops (Wareing *et al.*, 1986). This pathogen causes seedling blight, decay, root rot, reduction of seeds germination, plant populations, vigour and yield of surviving plants and produces also effects on others agricultural plant species like: *Oryza sativa L.* (Kumar *et al.*, 2011), *Solanum tuberosum L.* (Daami-Remadi *et al.*, 2008), *Vicia lens L.* (Chang *et al.*, 2008), *Brassica napus L.* (Hwang *et al.*, 2014), *Zea mays L.*, reducing root development and shoot growth (da Silva *et al.*, 2017) and for this reason is, along with other pathogens, the subject of several studies.

Research has shown significant variation in seedling vigour traits between different cultivars, populations and due to pathogens (Fakorede & Ojo, 1981; Kumar *et al.*, 2011; Xu *et al.*, 2019). Various methods have been developed to assess seedling vigor, including destructive and non-destructive measurements of emergence, growth rates, and biomass accumulation (Hu *et al.*, 2016).

In order to quantify the negative effects of pathogens or the beneficial effects of treatments applied in experiments of biocontrol, the germination rate dynamic is an investigated feature calculated with a manual count that is usually performed during the experiment with the aim of calculate the percentage of seeds germinated in precise moment (Kumar *et al.*, 2011).

Another useful features to evaluate seeds vitality is the seedling vigor which correlates with the speed of biomass development and thus with leaf area. Non-destructive measurement methods using orthogonal images have improved the accuracy of leaf area estimation in vegetable seedlings (Chien & Lin, 2005), while destructive methods involve determining the dry weight of the aerial part of plants (Anandan *et al.*, 2020).

Anyway, large-scale germination scoring experiments still commonly relies on human observation, so are labor intensive and prone to observer errors, constrained the frequency, scale and accuracy of such experiments leading to the necessity for automated methods (Colmer *et al.*, 2020).

Automated seed counting and germination vigor analysis have become increasingly important in agricultural research and seed testing. Several systems have been developed to address this need. SeedGerm combines hardware and software for automated seed imaging and machine learning-based phenotypic analysis across multiple crop species (Colmer *et al.*, 2020).

The aim of this work is to characterize the germination kinetics and the growth vigor of lettuce plants automatically using imaging techniques and low-cost camera. To introduce some variability in the germination phenomenon, half of the replicates were treated by adding a pathogen (*Rhizoctonia solani C.*) to the growth substrate, which is known in the literature (Aydin, 2022) to be capable of affecting the germinability and vigor of seedlings during the early vegetative stages. In this trial, an automated method will be proposed for counting the emerged seedlings, which will be repeated several times throughout the duration of the experiment. Additionally, at each count, the vigor of the seedlings will be evaluated by measuring the leaf area with top-view image.

## 2 Material and methods

### 2.1 Sample preparation

The soilborne fungal pathogen *Rhizoctonia solani* (Cooke) Wint, strain RS1, used in this study was isolated from millet (*Pennisetum glaucum L.*) kernels in 2012. The isolate was maintained on potato dextrose agar (PDA, Difco™) at 20°C and stored at 4°C. Three 1 cm$^2$ blocks of actively growing *Rhizoctonia solani* mycelium were added to a flask containing 200 g of pearl millet previously autoclaved twice at 121°C for 30 min with 100 ml of water. The inoculated flask was incubated for three weeks at 20°C. The grown mycelium was separated and placed at 30°C until completely dry and then inoculated into the soil at a concentration of 4 gkg$^{-1}$ one week before sowing. Six replicates (r) treated with the inoculum and an equal number of untreated controls were prepared. For each of 12 replicates, 49 baby lettuce seeds were sown, approximately 2 cm apart, in 100 g of substrate for sowing (Vigorplant) contained by 16 x 16 cm plastic supports. The tests lasted 10 days from the seeds sowing until harvesting.

### 2.2 Experimental setup

Lettuce plants were cultivated on a greenhouse at the Department of Agricultural and Environmental Sciences of the University of Milan (Italy) under controlled conditions at 24/20°C day/night temperature, with 60-75% relative humidity and 16 h photoperiod.

Color imaging was carried out by installing three TP-Link Tapo C-310 cameras (res. 1296 x 2304 pix.) above a workbench measuring 1 x 0.8 m. The cameras (*c*) were positioned at a distance of 0.8 m from the workbench, allowing for monitoring of the entire area with adequate resolution, each capturing images of four replicates simultaneously, for a total of 12 monitored units. These cameras were programmed via a dedicated application, scheduling five daily one-minute cycles of automatic video acquisition at 08:00, 11:00, 14:00, 17:00 and 20:00. The time considered in the experiment (*t*) is defined as the number of hours elapsed since the seeding moment.

This setup ensured consistent and parallel observation of treated and control groups under identical conditions.

### 2.3 Automatic counting validation

Count validation was performed manually at the end of the experiment and compared with the automated count obtained from the analysis of the final image acquired by each camera. This specific time point was selected for validation as it represents the most challenging stage for image-based quantification, due to the advanced development of seedlings and the formation of dense clusters that hinder accurate segmentation of individual plants.

### 2.4 System calibration

To enable the use of single video frames captured by the cameras, it was first necessary to correct for barrel distortion introduced by their lenses. This was accomplished through image rectification using the Camera Calibrator app available in MATLAB (R2023b; The MathWorks, Natick, Massachusetts). The calibration process involved acquiring 20 images of a dedicated checkerboard pattern, allowing the estimation of both intrinsic and extrinsic camera parameters. The calibration session, in addition to estimating the parameters and eliminate lens distortion, is also necessary to estimate the conversion parameter between pixels and corresponding surface area ($K_{conv}$) (~0.19 mm² pixel$^{-1}$), to assign the workspaces for the individual replicates and to determine the initial position of the centering marker.

To determine the spaces allocated to an individual replicates within the field of view of each camera, an image was acquired after placing the pot in their fixed positions occupied during the experiment and placing an AprilTags of the same size as the pot. AprilTags are graphical markers widely used in automation and robotics applications (Olson, 2011) to acquire information regarding the size, position and ID of the marker (in this case, from 1 to 4, corresponding to the replicates). The advantage of using these markers is that they can be automatically detected using specific functions (*readAprilTag* in MATLAB environment).

The calibration images, after being undistorted, are analyzed to search for the markers that identify the individual replicates. The vertices positions of the identified markers are used to generate corresponding polygons, which are converted using the *poly2mask* function into binary masks (*MaskR*) that are used during the analysis phase to segment the individual replicates, as illustrated in *Fig. 4*.

In addition, a centering marker (AprilTag ID 10, 30 × 30 mm) was installed at the center of the replicate area. This marker was used throughout the experiment to monitor its position and correct for minor camera displacements, ensuring precise spatial alignment of all acquired images. The marker serves a dual purpose: beyond spatial referencing, it also facilitates temporal alignment and assists in locating the gray reference patch present in each image. This reference is used to normalize the color channels by assigning it a fixed RGB value of [155, 155, 155] (Gray$_{(R,G,B)}$).

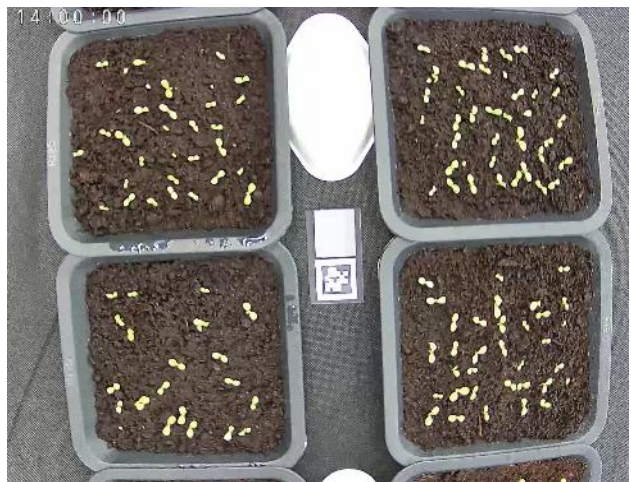

*Figure 1. Raw single frame extracted from the video stream of the camera. The image clearly shows pronounced barrel distortion. The four replicates, the central marker, and the gray reference are reported.*

Given the known dimensions of the markers, the metric conversion factor for each camera ($KConv_{(c)}$) was computed by dividing the actual marker area by the number of segmented pixels within the corresponding mask for each replicate. The resulting values were then averaged across replicates for each camera. This conversion factor is subsequently employed both in the distance-based clustering process described in *Sect.2.5.4* and in the estimation of leaf area for assessing seedling vegetative vigor, as detailed in *Sect.2.5.3*.

## 2.5 Data processing

### 2.5.1 Color normalization

The first step of the image analysis consists of normalizing the color channels. This was performed by sampling a 10 × 10 pix. region of interest (ROI) from the central area of the gray reference to compute the mean intensity values for the red, green, and blue channels ($Mean_{(R,G,B)}$). These values were then compared to the predefined reference gray value ($Gray_{(R,G,B)}$) to derive a normalization coefficient for each channel. The resulting coefficients, which reflect the lighting conditions at the time of image acquisition, were applied to normalize the color channels across different time points, as described in *Eq. 1*.

$$INorm_{(R,G,B)} = \frac{Gray_{(R,G,B)}}{Mean_{(R,G,B)}} \cdot I_{(R,G,B)} \quad [Eq.1]$$

### 2.5.2 Image analisys

Following color normalization, the next phase of image analysis involves plant segmentation, assignment of each plant to its corresponding replicate, and clustering of elements within each replicate based on proximity to generate individual polygons. The images analyzed are of limited quality, as they are extracted from single frames of highly compressed video streams recorded by low-cost cameras. Therefore, several preprocessing steps are required to enhance segmentation accuracy.

The first preprocessing step focuses on improving edge definition. To this end, the *imsharpen* function in MATLAB is employed, which utilizes the unsharp masking technique to enhance contrast along color boundaries. This method sharpens the image by subtracting a blurred version from the original, thereby emphasizing edges and fine details.

Following the sharpening, image edges are more accurately detected by computing the gradient using the Sobel method (*imgradient*). In this process, only the green channel is analyzed, as it provides the most relevant edge information for plant structures. A threshold is then applied to the gradient matrix, determined using Otsu's method, to exclude pixels with gradient values above this threshold. This approach enables the generation of a binary mask in which distinct blobs correspond to regions of relatively uniform green coloration. To further refine the result and reduce computational load in subsequent steps, a minimum size filter of 25 pix. (equivalent to 4.75 mm²) is applied, removing less significant elements or to remove some background elements erroneously segmented.

The next step of the process involves segmenting only the areas of predominantly green color. This was achieved by calculating the excess-green vegetation index (*ExG*) (see *Eq. 2*) at the pixel level and imposing a minimum threshold that a pixel must meet to be classified as green and, therefore, as plant tissue. Typically, the segmentation threshold is set at the pixel level, but in this case, to enhance the accuracy of the segmentation process, the threshold was applied at the average value for each individual blob. The threshold value was determined using Otsu's method ($T_{ExG} = 25$) on a sample of images and was applied during the analysis to all the images in the experimental database. The threshold value used aligns with other examples found in the literature, such as for the Canopeo, a tool useful for automatic segmentation of vegetation pixels in an image (Patrignani & Ochsner, 2015).

$$ExG_{(c,t)} = 2\,g - r - b \quad [Eq.2]$$

$$GreenP_{(c,r,t)} = (ExG_{(c,t)} > T_{ExG})\,\&\,MaskR_{(r)} \quad [Eq.3]$$

### 2.5.3 Growth vigor

For each camera (*c*) the estimation of the vegetative vigor of the individual replicates (*r*), defined as *LeafArea*$_{(c,r,t)}$, is based on estimating the leaf area measurement. This value was obtained by multiplying the spatial conversion coefficient of each camera ($K_{Conv(c)}$) by the sum of the segmented predominantly green pixels in each of the replicates (*GreenP*$_{(c,r,t)}$) at a certain moment in the experiment (*t*), as outlined in *Eq. 4*.

$$LeafArea_{(c,r,t)} = K_{Conv(c)} \cdot \sum GreenP_{(c,r,t)} \quad [mm^2] \quad [Eq.\,4]$$

### 2.5.4 Clusterization and counting

The output of the segmentation step serves as the foundation for identifying and counting individual plants. For each replicate, all segmented green regions (*GreenP*$_{(c,r,t)}$) from a given acquisition—representing a specific time point after seeding—are considered. A clustering algorithm is then applied, grouping blobs based on a spatial proximity criterion: blobs whose maximum inter-pixel distance is less than 5 mm are assigned to the same cluster. This procedure enables the definition of cluster boundaries, which are then used to generate corresponding polygons (*Polyg*$_{(c,r,t)}$) via MATLAB's *polyshape* function, which constructs two-dimensional polygons from vertex data.

The estimation of germinated seedlings at a given time point is performed by analysing these polygons in a reverse temporal sequence. Starting from the latest acquisition, each polygon is compared with those from earlier time points. A polygon is classified as a new emergence event if it does not spatially overlap with any polygon detected in preceding images. These non-overlapping polygons, defined as emergence polygons (*PE*$_{(c,r,t)}$), are used to identify the timing of seedling emergence and subsequently characterize germination dynamics.

```
Input: Polyg(c,r,t)
for t=N(acquisition):-1:1
 for r=1:N(replicates)
  for c=1:N(cameras)
   for ts=t:-1:1
    check overlay between Polyg(c,r,t) and Polyg(c,r,ts)
    if Polyg(c,r,t) has no overlay
     PE(c,r,t)= Polyg(c,r,t)
    end
   end
  end
 end
end
```

*Figure 2. Pseudocode summarizing the main steps involved in the automated counting of germinated seedlings.*

This backward-tracking approach is particularly effective in the later stages of the experiment, when growing seedlings tend to overlap, making it challenging to accurately delineate individual boundaries through image segmentation alone.

Automated counts at the final time point were validated by comparing them with manual counts performed at the end of the experiment, demonstrating the reliability of the proposed method.

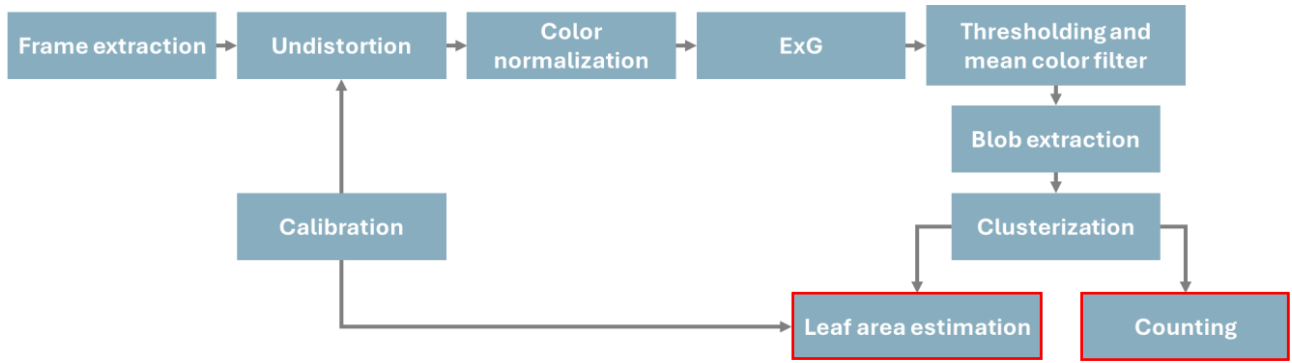

*Figure 3. Flowchart of the image analysis and germination counting workflow. The diagram outlines the main processing steps performed by the algorithm, from image acquisition and preprocessing (including distortion correction and color normalization), to plant segmentation, clustering, and polygon extraction. The final stages involve temporal comparison across image sequences to identify seedling emergence events and compute germination counts over time.*

## 3 Results

### 3.1 Calibration

The calibration process produced 4 binary masks for each camera, corresponding to the individual replicates. As shown in *Fig. 4b*, the masks derived from a single camera are displayed using different colors for improved visual clarity. Beyond delineating the replicate regions, this process also allowed for the estimation of the metric conversion factor, *KConv$_{(c)}$*, with the results summarized in *Tab. 1*. The computed coefficients vary by approximately 0.1%, a difference attributed to slight discrepancies in the elevation and alignment of the cameras.

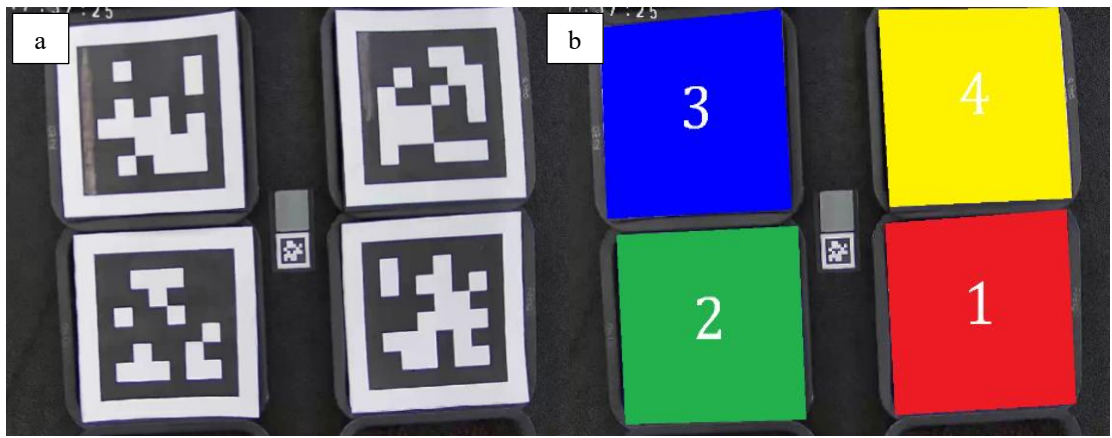

*Figure 4. Example of the calibration process used to define the area assigned to each replicate within the image and to estimate the size of a single pixel. On the left, the calibration image displays the AprilTags used to delimit individual replicates; on the right, the overlay shows the corresponding binary masks in red, green, blue, and yellow*

*Table 1. Linear dimensions and the corresponding pixel area for each camera*

|  | Camera | | |
|---|---|---|---|
|  | 1 | 2 | 3 |

| pixel area [mm²pix⁻¹] | 0.195 | 0.194 | 0.194 |
|---|---|---|---|
| pixel side length [mm] | 0.441 | 0.440 | 0.440 |

## 3.2 Image segmentation

The segmentation workflow proved to be effective for the experimental objectives. *Fig. 5* illustrates a representative output from each step of the process. In the first stage (*Fig. 5a*), the undistortion of the raw frame corrected the barrel distortion and restored the proper geometry of the scene.

The gradient-based filter successfully removed transitional areas where the green channel gradient exceeded the threshold, while preserving critical plant structures, particularly the leaves (*Fig. 5b*). In some cases, the central region of a seedling was excluded due to its narrowness and its surrounding background, which led it to be misclassified as a transition zone (*Fig. 5c*). Despite this, the final plant segmentation remained robust thanks to the 5mm distance threshold applied during clustering.

Lastly, the classification of the segmented polygons for each replicate—represented in different colors—is shown in *Fig. 5d*. These polygons correspond to the output of the analysis performed on a single image frame. The counting process can be initiated from the most recent output and proceeds backward in time, iteratively considering the results from previous image acquisitions back to the first frame, allowing the identification of the moment of seedling emergence.

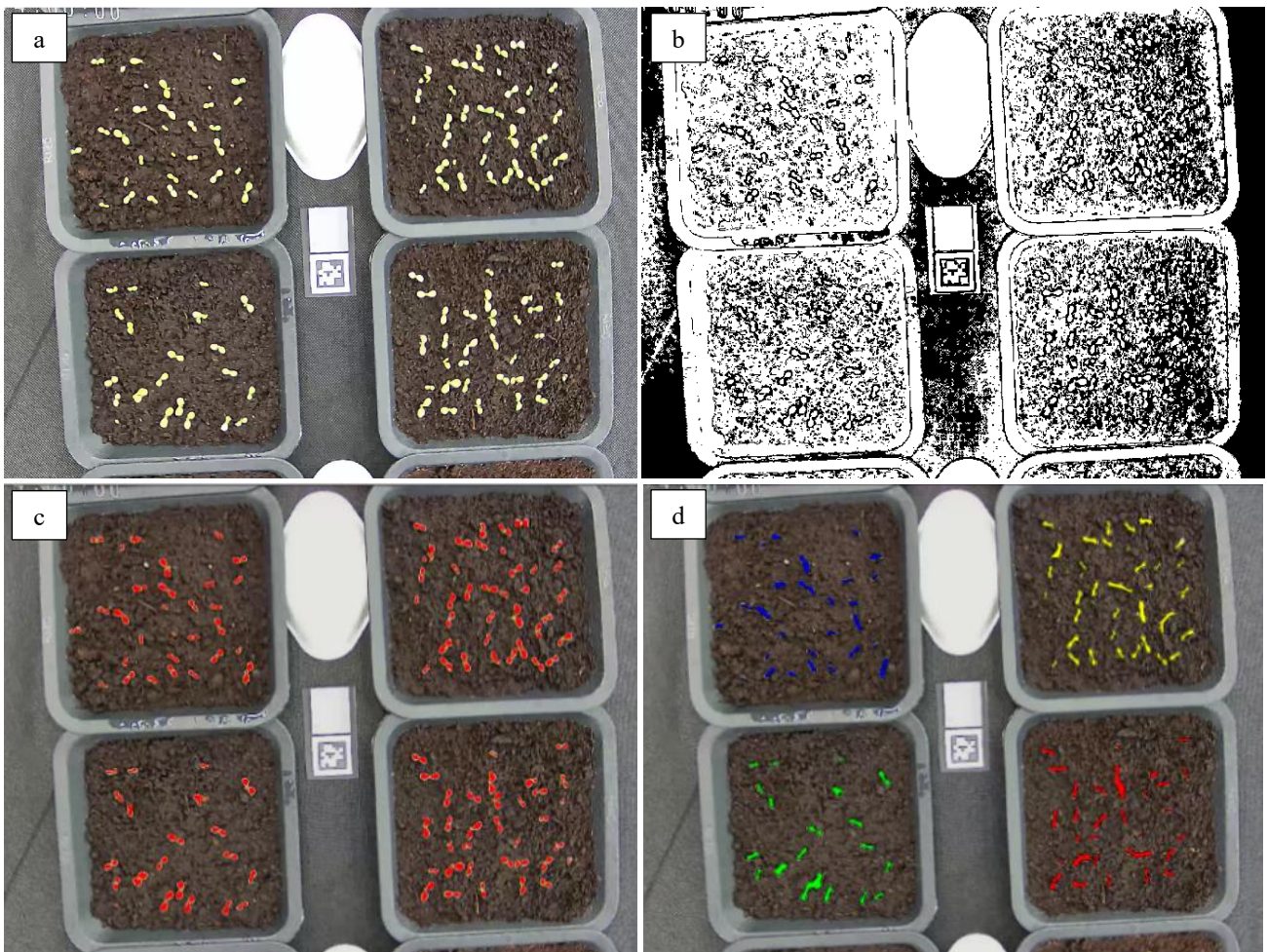

*Figure 5. Segmentation workflow applied to a single video frame. (a) Rectified image after undistortion, correcting for lens distortion and restoring geometric accuracy. (b) Gradient-based filtering removes high-contrast transitions while preserving seedling features, especially leaf structures. (c) Example of blob segmentation. (d) Result of the classification of segmented clusters and corresponding polygons, with each replicate area represented using a distinct color for clarity.*

### 3.3 Clusterization and counting

The clustering approach proved effective in restoring the structural integrity of the segmented regions and accurately isolating individual seedlings. However, as seedling size increased during later stages of development, clustering occasionally grouped multiple individuals into a single polygon. This limitation was illustrated in *Fig. 5d*, where polygons, color-coded by replicate ID, were generated using the boundaries function.

The proposed counting strategy—based on identifying the emergence polygons and tracing their first appearance—demonstrated high accuracy. Temporal alignment of the image sequences was also successful, as shown in *Fig. 6b*, which presents a stacked view of overlapping polygons across acquisition epochs. Early-stage polygons appear on top, representing the initial emergence events, while those from later stages become progressively merged due to increased plant growth and contact among seedlings. This layering also confirms the spatial stability achieved through the centering marker.

The effectiveness of the method was validated by comparing the final automated counts with manual counts performed at the end of the experiment, which included 294 seeds. As shown in *Fig. 6a*, the results yielded an RMSE of 1.12 and a determination coefficient ($R^2$) of 0.98. The final time point was selected for validation as it represents both the standard endpoint in manual assessments and the most complex phase for image-based segmentation due to extensive overlap among seedlings.

As expected, the germination rate of the untreated seeds was higher. Moreover, the intra-treatment variability among replicates was significantly lower compared to the inoculated group, indicating a more uniform response in the absence of the pathogen.

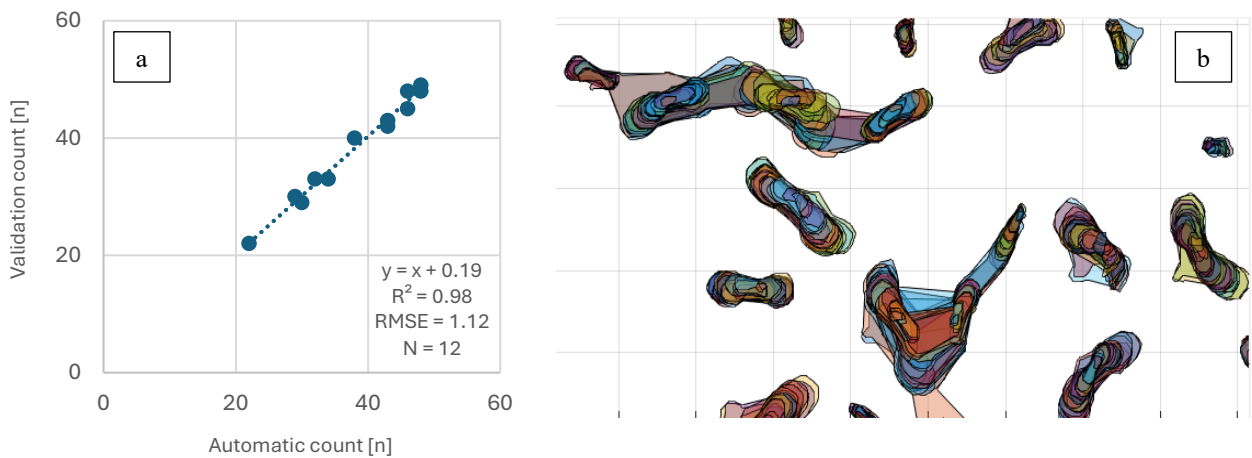

*Figure 6. (a) Validation results of the seedling counting method: the scatter plot shows the correlation between the automated and manual counts, along with the regression line, R² value, and RMSE. (b) Temporal overlay of the polygons extracted from a portion of a replicate during the experiment. The polygons are stacked chronologically, with the uppermost representing the earliest detections and the lower ones corresponding to later time points. The image highlights the high accuracy of the temporal alignment achieved throughout the monitoring process.*

### 3.4 Growth vigour and counting

The estimation of seedling vigor, illustrated in *Fig. 7b*, reveals periodic fluctuations in the vigor curves, marked by alternating peaks. These patterns are likely associated with diurnal leaf movements, such as variations in leaf inclination and partial closure during nighttime hours.

Data analysis revealed that the untreated replicates consistently exhibited higher average vigor compared to those inoculated with the pathogen, in line with the initial expectations. Notably, the inoculated replicates showed the lowest vigor values among all treatments, further confirming the detrimental impact of the pathogen on seedling development.

The germination data further support the impact of the pathogen, as inoculated seeds exhibited reduced germination rates and increased variability among replicates compared to the untreated controls. This confirm—as expected—that *Rhizoctonia* infection not only diminishes seed viability but also leads to inconsistent germination outcomes across experimental units.

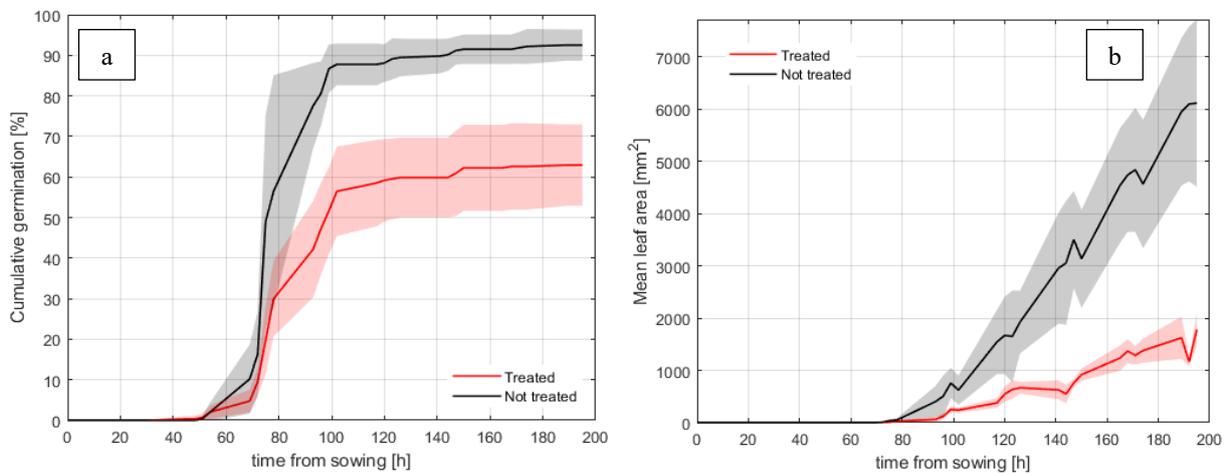

*Figure 7. (a) Time-course plot showing the number of germinated seedlings throughout the experiment, illustrating the emergence dynamics. The shaded area around the line represents the variability as ±1 standard deviation. (b) Biomass development dynamics over time, with the shaded region indicating ±1 standard deviation around the mean values.*

## 4 Conclusions and discussion

Overall, the dynamics confirm the findings observed in the analysis of seedling vigor. The treated replicates not only exhibited lower vigor and germinability levels but also showed greater variability. The treated replicates are characterized by different levels of seedling emergence, with increased variability in this phenomenon. Regarding vegetative vigor, the pathogen had a significant negative effect, reducing vigor by 70% by the end of the experiment, substantially limiting the vigor levels achieved by the emerged seedlings.

Other comparable systems have achieved results in line with those obtained in this study. For instance, SeedGerm (Colmer et al., 2020) reported a similar determination coefficient for automatic seedling counting, despite employing higher-resolution cameras—5 MP and 8 MP HD USB cameras in its two configurations—compared to the 2.7 MP cameras used here. Additionally, the imaging setup in the present experiment involves a greater camera-to-subject distance, resulting in a significantly lower spatial resolution. A noteworthy difference lies in the experimental conditions: in this study, seeds were sown in a cultivation substrate, which introduces additional complexity to the segmentation and counting process. In contrast, other approaches often place seeds on plates with background colors selected to maximize contrast, thereby simplifying image analysis. However, such idealized setups may not accurately reflect real-world scenarios, where seeds interact with heterogeneous substrates and potential biocontrol agents. The conditions adopted in this study offer a more realistic context for evaluating early plant development under biologically relevant conditions

In comparison with the approach presented by Fuentes-Peñailillo (2023) where a deep-learning application based on SSD MobileNet was employed for crop detection in seedling trays, the method proposed in this study offers a different and complementary perspective. While their system allows for instant seedling detection through neural networks—achieving detection accuracies ranging from 57% to 96% depending on crop type and tray layout—it also highlights significant limitations related to environmental variability and dataset heterogeneity. These factors notably affected the generalizability of the model, underscoring the importance of both experimental setup and classification robustness.

Conversely, the image-based pipeline introduced in this study does not rely on deep learning models for instantaneous classification, but instead focuses on temporal tracking of plant emergence and growth. By characterizing the germination kinetics and biomass development with high temporal resolution, this method allows for a more detailed and accurate analysis of seedling dynamics. Moreover, the consistent alignment and normalization steps contribute to minimizing variability across replicates, thereby enhancing the reliability of automatic measurements. While real-time prediction is not the primary goal, the increased precision and interpretability of results make this approach particularly suitable for controlled experiments requiring high data granularity.

The method for estimating the count of emerged seedlings and their vigour proved suitable for achieving the objectives of the experiment, providing excellent results in evaluating the kinetics of the observed variables. This method could be employed in more complex experiments where it is necessary to assess the effects of multiple treatments and accurately characterize the response of the individuals. In this case, all the collected images were used without discarding any.

Future developments could involve improving camera quality without increasing costs, allowing for higher image resolution and especially avoiding the use of highly compressed frames that degrade image quality, making analysis less efficient and less accurate.

# Reference

[1] Anandan, A., Mahender, A., Sah, R. P., Bose, L. K., Subudhi, H., Meher, J., Reddy, J. N., & Ali, J. (2020). Non-destructive phenotyping for early seedling vigor in direct-seeded rice. Plant Methods, 16(1). https://doi.org/10.1186/s13007-020-00666-6

[2] Aydin, M. H. (2022). Rhizoctonia solani and Its Biological Control. In Türkiye Tarımsal Araştırmalar Dergisi (Vol. 9, Issue 1, pp. 118–135). Turkish Journal of Agricultural Research (TUTAD). https://doi.org/10.19159/tutad.1004550

[3] Chang, K. F., Hwang, S. F., Gossen, B. D., Turnbull, G. D., Wang, H., & Howard, R. J. (2008). Effects of inoculum density, temperature, seeding depth, seeding date and fungicidal seed treatment on the impact of Rhizoctonia solani on lentil. In Canadian Journal of Plant Science (Vol. 88, Issue 4, pp. 799–809). Canadian Science Publishing. https://doi.org/10.4141/p06-020

[4] Chien, C. F. & Lin, T. T. (2005). Non-destructive growth measurement of selected vegetable seedlings using orthogonal images. In Transactions of the ASAE (Vol. 48, Issue 5, pp. 1953–1961). American Society of Agricultural and Biological Engineers (ASABE). https://doi.org/10.13031/2013.19987

[5] Colmer, J., O'Neill, C. M., Wells, R., Bostrom, A., Reynolds, D., Websdale, D., Shiralagi, G., Lu, W., Lou, Q., Le Cornu, T., Ball, J., Renema, J., Flores Andaluz, G., Benjamins, R., Penfield, S., & Zhou, J. (2020). SeedGerm: a cost-effective phenotyping platform

for automated seed imaging and machine-learning based phenotypic analysis of crop seed germination. In New Phytologist (Vol. 228, Issue 2, pp. 778–793). Wiley. https://doi.org/10.1111/nph.16736

[6] Daami-Remadi, M., Zammouri, S., & El Mahjoub, M. (2008). Effect of the level of seed tuber infection by Rhizoctonia solani at planting on potato growth and disease severity. The African Journal of Plant Science and Biotechnology, 2(1), 34-38.

[7] da Silva, M. P., Tylka, G. L., & Munkvold, G. P. (2017). Seed Treatment Effects on Maize Seedlings Coinfected with Rhizoctonia solani and Pratylenchus penetrans. In Plant Disease (Vol. 101, Issue 6, pp. 957–963). Scientific Societies. https://doi.org/10.1094/pdis-10-16-1417-re

[8] Fakorede, M. A. B., & Ojo, D. K. (1981). Variability for Seedling Vigour in Maize. In Experimental Agriculture (Vol. 17, Issue 2, pp. 195–201). Cambridge University Press (CUP). https://doi.org/10.1017/s0014479700011455

[9] Fuentes-Peñailillo, F., Carrasco Silva, G., Pérez Guzmán, R., Burgos, I., & Ewertz, F. (2023). Automating Seedling Counts in Horticulture Using Computer Vision and AI. Horticulturae, 9(10), 1134. https://doi.org/10.3390/horticulturae9101134

[10] Gonzalez Garcia, V., Portal Onco, M. A., & Rubio Susan, V. (2006). Review. Biology and systematics of the form genus Rhizoctonia. Spanish Journal of Agricultural Research, 4(1), 55–79. https://doi.org/10.5424/sjar/2006041-178[] Gullino, M.L., Gilardi, G., Garibaldi, A. (2019). Ready-to-Eat Salad Crops: A Plant Pathogen's Heaven. Plant disease, 103 (9): 2153-2170. https://doi.org/10.1094/PDIS-03-19-0472-FE

[11] Hu, Q., Fu, Y., Guan, Y., Lin, C., Cao, D., Hu, W., Sheteiwy, M., & Hu, J. (2016). Inhibitory effect of chemical combinations on seed germination and pre-harvest sprouting in hybrid rice. In Plant Growth Regulation (Vol. 80, Issue 3, pp. 281–289). Springer Science and Business Media LLC. https://doi.org/10.1007/s10725-016-0165-z

[12] Hwang, S. F., Ahmed, H. U., Turnbull, G. D., Gossen, B. D., & Strelkov, S. E. (2014). The effect of seed size, seed treatment, seeding date and depth on Rhizoctonia seedling blight of canola. In Canadian Journal of Plant Science (Vol. 94, Issue 2, pp. 311–321). Canadian Science Publishing. https://doi.org/10.4141/cjps2013-294

[13] Kumar, K. V. K., Reddy, M. S., Kloepper, J. W., Lawrence, K. S., Yellareddygari, S. K. R., Zhou, X. G., ... & Miller, M. E. (2011). Screening and selection of elite plant growth promoting rhizobacteria (PGPR) for suppression of Rhizoctonia solani and enhancement of rice seedling vigor. J. Pure Appl. Microbiol, 5(2), 1-11.

[14] Losio, M.N., Pavoni, E., Bilei, S., Bertasi, B., Bove, D., Capuano, F., Farneti, S., Blasi, G., Comin, D., Cardamone, C., Decastelli, L., Delibato, E., De Santis, P., Di Pasquale, S., Gattuso, A., Goffredo, E., Fadda, A., Pisanu, M., De Medici, D. (2015). Microbiological survey of raw and ready-to-eat leafy green vegetables marketed in Italy. International Journal of Food Microbiology, 210: 88-91, https://doi.org/10.1016/j.ijfoodmicro.2015.05.026

[15] Olson, E. (2011). AprilTag: A robust and flexible visual fiducial system. In 2011 IEEE International Conference on Robotics and Automation. 2011 IEEE International Conference on Robotics and Automation (ICRA). IEEE. https://doi.org/10.1109/icra.2011.5979561

[16] Patrignani, A., & Ochsner, T. E. (2015). Canopeo: A Powerful New Tool for Measuring Fractional Green Canopy Cover. In Agronomy Journal (Vol. 107, Issue 6, pp. 2312–2320). Wiley. https://doi.org/10.2134/agronj15.0150

[17] Raid, R.N., Sandoya-Miranda, G. (2024). Diseases of Lettuce. In: Elmer, W.H., McGrath, M., McGovern, R.J. (eds) Handbook of Vegetable and Herb Diseases. Handbook of Plant Disease Management. Springer, Cham. https://doi.org/10.1007/978-3-030-35512-8_52-1

[18] Tomasi N., Pinton R., Dalla Costa L., Cortella G., Terzano R., Mimmo T., Scampicchio M., Cesco S. (2015). New 'solutions' for floating cultivation system of ready-to-eat salad: A review. Trends in Food Science & Technology, 46: 267-276, https://doi.org/10.1016/j.tifs.2015.08.004

[19] Verwaaijen, B., Wibberg, D., Kröber, M., Winkler, A., Zrenner, R., Bednarz, H., Niehaus, K., Grosch, R., Pühler, A., & Schlüter, A. (2017). The Rhizoctonia solani AG1-IB (isolate 7/3/14) transcriptome during interaction with the host plant lettuce (Lactuca sativa L.). In L. S. van Overbeek (Ed.), PLOS ONE (Vol. 12, Issue 5, p. e0177278). Public Library of Science (PLoS). https://doi.org/10.1371/journal.pone.0177278

[20] Wareing, P. W., Wang, Z., Coley-Smith, J. R., & Jeves, T. M. (1986). Fungal pathogens in rotted basal leaves of lettuce in Humberside and Lancashire with particular reference to Rhizoctonia solani. Plant Pathology, 35(3), 390–395. https://doi.org/10.1111/j.1365-3059.1986.tb02031.x

[21] Xu, L., Guo, L., You, H., Zhang, O., & Xiang, X. (2019). Novel haplotype combinations reveal enhanced seedling vigor traits in rice that can accurately predict dry biomass accumulation in seedlings. In Breeding Science (Vol. 69, Issue 4, pp. 651–657). Japanese Society of Breeding. https://doi.org/10.1270/jsbbs.19087